\documentclass[a4paper]{article}

\usepackage{INTERSPEECH2018}

\usepackage{cite,url}
\usepackage{tabularx}

\title{Acoustic and Textual Data Augmentation for Improved ASR of Code-Switching Speech}
\name{Emre Y\i lmaz$^{1,2}$, Henk van den Heuvel$^1$ and David A. van Leeuwen$^1$\thanks{This research is funded by the NWO Project 314-99-119 (Frisian Audio Mining Enterprise).}}
\address{
  $^1$ CLS/CLST, Radboud University, Nijmegen, Netherlands \\
  $^2$ Dept. of Electrical and Computer Engineering, National University of Singapore, Singapore}
\email{\{e.yilmaz, h.vandenheuvel, d.vanleeuwen\}@let.ru.nl}

\begin{document}

\maketitle
\begin{abstract}
In this paper, we describe several techniques for improving the acoustic and language model of an automatic speech recognition (ASR) system operating on code-switching (CS) speech. We focus on the recognition of Frisian-Dutch radio broadcasts where one of the mixed languages, namely Frisian, is an under-resourced language. In previous work, we have proposed several automatic transcription strategies for CS speech to increase the amount of available training speech data. In this work, we explore how the acoustic modeling (AM) can benefit from monolingual speech data belonging to the high-resourced mixed language. For this purpose, we train state-of-the-art AMs, which were ineffective due to lack of training data, on a significantly increased amount of CS speech and monolingual Dutch speech. Moreover, we improve the language model (LM) by creating code-switching text, which is in practice almost non-existent, by (1) generating text using recurrent LMs trained on the transcriptions of the training CS speech data, (2) adding the transcriptions of the automatically transcribed CS speech data and (3) translating Dutch text extracted from the transcriptions of a large Dutch speech corpora. We report significantly improved CS ASR performance due to the increase in the acoustic and textual training data.
     
\end{abstract}
\noindent\textbf{Index Terms}:  code-switching, bilingual ASR, under-resourced languages, Frisian language

\section{Introduction}

Impact of CS and other kinds of language switches on the speech-to-text systems have recently received research interest, resulting in several robust acoustic modeling~\cite{stemmer2001,lyu2006,vu2012,modipa2013,lyudovyk2014,wu2014,yilmaz2016_2,weiner2012,lyu2013,yeong2014,mabokela2014} and language modeling~\cite{li2012,adel2013,adel2014,zeng2017,hamed2017,westhuizen2017} approaches for CS speech. One fundamental approach is to label speech frames with the spoken language and perform recognition of each language separately using a monolingual ASR system at the back-end~\cite{weiner2012,lyu2013,yeong2014,mabokela2014}. These systems have the tendency to suffer from error propagation between the language identification front-end and ASR back-end, since language identification is still a challenging problem especially in case of intra-sentence CS. To alleviate this problem, all-in-one ASR approaches, which do not directly incorporate a language identification system, have also been proposed~\cite{lyu2006,lyudovyk2014,yilmaz2016_2}. Our research in the FAME! project focuses on developing an all-in-one CS ASR system using a Frisian-Dutch bilingual acoustic and language model that allows language switches.~\cite{yilmaz2016_2,yilmaz2016_4}.

One of the bottlenecks for building a CS ASR system is the lack of training speech and text data to train reliable acoustic and language models that can accurately recognize the uttered word and its language. The latter is relevant in language pairs such as Frisian and Dutch with orthographic similarity and shared vocabulary. For this purpose, we have proposed automatic transcription strategies developed for CS speech to increase the amount of training speech data in previous work~\cite{yilmaz2017_2, yilmaz2018}. 

On account of the increased training data \cite{yilmaz2018}, we reported improvements in the ASR and CS detection accuracy using fully connected deep neural networks (DNN). In this work, we investigate how to combine the out-of-domain speech data from the high-resourced mixed language with this increased amount of CS speech data for acoustic training of a more recently proposed neural network architecture consisting of time-delay and recurrent layers~\cite{peddinti2017}. In pilot experiments, these more sophisticated models have been found to provide worse performance due to lack of training data. We further address the challenge that Dutch spoken in the target speech differs from standard Dutch due to a noticeable Frisian accent. For this purpose, we include training Flemish data which is a language variety spoken in Belgium with the aim of making the final acoustic models more accent-robust.

For the language modeling, we have thus far incorporated a standard bilingual language model which is mostly trained on a combination of monolingual text from Frisian and Dutch. The only component in the training text with CS is the transcriptions of the FAME! Corpus~\cite{yilmaz2016} training speech data. Motivated by the success of automatic training data generation for acoustic modeling, we generate code-switching text from transcriptions of the spoken data which has been found to have lower perplexity on the development data compared to text from other written resources.

CS text is generated either by training long short-term memory (LSTM) language models on the very small amount of CS text extracted from the transcriptions of the training speech data and synthesize much larger amounts of CS text using these models~\cite{sutskever2011} or by translating Dutch text extracted from the transcriptions of a large Dutch speech corpora. The latter provides CS text as some Dutch words, such as proper nouns (most commonly person, place and institution names) which are a prominent source of code-switching, are not translated to Frisian while the neighboring words are. We further include the transcriptions provided by the best performing automatic transcription strategies described in~\cite{yilmaz2018}.


\vspace{-0.2cm}
\section{Frisian-Dutch Radio Broadcast Database}
\label{sec:database}

The bilingual FAME! speech database, which has been collected in the scope of the \textit{Frisian Audio Mining Enterprise} project, contains radio broadcasts in Frisian and Dutch. The Frisian language shows many parallels with Old English. However, nowadays it is under growing influence of Dutch due to long lasting and intense language contact. Frisian has about half a million speakers. A recent study shows that about 55\% of all inhabitants of Frysl\^{a}n speak Frisian as their first language, which is about 330,000 people~\cite{provinciefryslan}. All speakers of Frisian are at least bilingual, since Dutch is the main language used in education in Frysl\^{a}n.

The FAME! project aims to build a spoken document retrieval system operating on the bilingual archive of the regional public broadcaster Omrop Frysl\^{a}n (Frisian Broadcast Organization). This bilingual data contains Frisian-only and Dutch-only utterances as well as mixed utterances with inter-sentential, intra-sentential and intra-word CS~\cite{myers1989}. To be able to design an ASR system that can handle the language switches, a representative subset of recordings has been extracted from this radio broadcast archive. These recordings include language switching cases and speaker diversity, and have a large time span (1966--2015).

The radio broadcast recordings have been manually annotated and cross-checked by two bilingual native Frisian speakers. The annotation protocol designed for this CS data includes three kinds of information: the orthographic transcription containing the uttered words, speaker details such as the gender, dialect, name (if known) and spoken language information. The language switches are marked with the label of the switched language. For further details, we refer the reader to~\cite{yilmaz2016}. As of February 2017, the standard FAME training setup is integrated in Kaldi toolkit~\cite{kaldi} as one of the example recipes.
\vspace{-0.2cm}
\section{Acoustic Modeling}
\label{sec:am}

In previous work, we described several automatic annotation approaches to enable using of a large amount of raw bilingual broadcast data for acoustic model training in a semi-supervised setting. For this purpose, we performed various tasks such as speaker diarization, language and speaker recognition and LM rescoring on raw broadcast data for automatic speaker and language tagging~\cite{yilmaz2017_2, yilmaz2018} and later used this data for acoustic training together with the manually annotated (reference) data. These approaches improved the recognition performance on the CS due to the significant increase in the available CS training data.

This work further focuses on the possible improvements in the acoustic modeling that can be obtained using other datasets with a much larger amount of monolingual speech data from the high-resourced mixed language, which is Dutch in our scenario. Previously, adding even a portion of this Dutch data resulted in severe recognition accuracy loss in the low-resourced mixed language due to the data imbalance between the mixed languages in the training data. Therefore, only a small portion of the available Dutch data could be used together with the Frisian data. 

After the significant increase in the CS training speech data, we explore to what extend one can benefit from the greater availability of resources for Dutch. Given that the Dutch spoken in the target CS context is characterized by the West Frisian accent~\cite{bezooijen1999}, we further include speech data from a language variety of Dutch, namely Flemish, to investigate its contribution towards the accent-robustness of the final acoustic model.
\vspace{-0.2cm}
\section{Language Modeling}
\label{sec:lm}

Language modeling in a CS scenario mainly suffers from lack of adequate training material, as CS rarely occurs (mostly in an informal context such personal messages and tweets) in written resources. Therefore, finding enough training material for a language model that can model word sequences with CS is very challenging. In our previous experiments, our main source for CS text was the transcriptions of the training speech data which comprises a very small proportion of the bilingual training text compared to the monolingual Frisian and monolingual Dutch text.

Text generation using recurrent neural network (RNN) architectures is a common application that can be used to address this problem. Creating text of similar nature to the available limited amount of CS text data in one straightforward way of remedying this imbalance in the bilingual training text corpora. For this purpose, we train an LSTM-based language model on the transcriptions of the training CS speech data from the FAME! Corpus and generate CS text to investigate if including various amount of generated CS text in the training text corpus reduces the perplexity on the transcriptions of the development and test speech data.

Language models trained on machine translated text has been found to be useful in a monolingual low-resourced setting~\cite{jensson2009}. We apply the same idea under a bilingual scenario with CS speech. As the source text, we use the transcriptions of the speech data belonging to Dutch as the high-resourced mixed language. Using machine translated text is expected to improve the CS language model in two ways: (1) creating CS examples in the presence of proper nouns such as institution names, person names, place names in Dutch and (2) generating CS word sequences that are extracted from a spoken corpus which is much larger than the FAME! Corpus. It is important to note that the quality of the translation highly depends on the language models used for the machine translation task. In this work, we used an open-source webservice\footnote{https://taalweb.frl/oersetter} for our Frisian-Dutch machine translation system which uses different language models\footnote{https://github.com/proycon/oersetter-models} from the baseline bilingual model used in this work.

As a third source of automatically generated CS text, we use the output of the automatic transcription strategies proposed for CS speech~\cite{yilmaz2018}. These automatic transcriptions are created using either a bilingual ASR system (bilingual strategies) or two monolingual ASR systems (monolingual strategies) based on the preprocessing applied to the raw broadcast recordings. The corresponding bilingual and monolingual LMs are trained on the baseline bilingual text corpus. Given that these automatic transcriptions are the mostly likely word sequence hypothesized based on both the acoustic and language model scores, they potentially contain new word sequences with CS that are unknown to the baseline LM. Adding this hypothesized text to the LM training corpus is expected to provide enriched LMs with new CS examples.

\begin{table}
\centering
\addtolength{\tabcolsep}{-4.2pt}
\caption{Acoustic data composition of different training setups used in the recognition experiments (in hours)}
\vspace{-0.15cm}
\begin{tabular}{| l | c | c | c | c | c |} 
\hline
Training data & Annot. & Frisian & Dutch & Flemish & Total  \\
\hline\hline
(1) FAME!~\cite{yilmaz2016} & Manual &   8.5   &   3.0  &    -    & 11.5  \\
\hline
(2) Frisian Broad.~\cite{yilmaz2018} & Auto. &   \multicolumn{2}{c|}{125.5}  & - & 125.5  \\
\hline
(3) CGN-NL~\cite{cgn} & Manual & - & 442.5 & - & 442.5 \\
\hline
(4) CGN-VL~\cite{cgn} & Manual & - & - & 307.5 & 307.5 \\
\hline
\end{tabular}
\vspace{-0.7cm}
\label{tab:data}
\end{table}

\begin{table*}
\centering
\caption{Perplexities obtained on the development and test transcriptions - The total number of words in each component is given in parenthesis. (AA: Automatic annotation, MT: Machine translated)}
\addtolength{\tabcolsep}{-2pt}
\vspace{-0.2cm}
\begin{tabular}{| l | l | c | c |}
\hline
LM        	&	Training text corpus &  Devel & Test \\
\hline \hline
(1) Baseline LM  &   Orig. (46M) & 257.9	 &	231.6    \\
\hline \hline
(2) LM\_GEN-10M &   Orig. (46M) + Text Gen. (10M) & 201.4 & 179.4\\
\hline
(3) LM\_GEN-25M &   Orig. (46M) + Text Gen. (25M) & 195.4 & 173.0\\
\hline
(4) LM\_GEN-50M &   Orig. (46M) + Text Gen. (50M) & 195.2 & 173.2\\
\hline
(5) LM\_GEN-75M &   Orig. (46M) + Text Gen. (75M) & 197.3 & 175.1\\
\hline \hline
(6) LM\_GEN-50M\_AA-M &  Orig. (46M) + Text Gen. (50M) + AA-Monoling. (1.5M) & 188.0 & 165.7\\
\hline
(7) LM\_GEN-50M\_AA-B &  Orig. (46M) + Text Gen. (50M) + AA-Biling. (1.5M) & 187.8 & 166.5\\
\hline
(8) LM\_GEN-50M\_AA-MB &  Orig. (46M) + Text Gen. (50M) + AA-Both (3M) & 179.8 & 157.9 \\
\hline \hline
(9) LM\_GEN-50M\_AA-MB\_MT\ &  Orig. (46M) + Text Gen. (50M) + AA-Both (3M) + MTranslated (8.5M) & \bf{178.1} & \bf{155.8} \\
\hline 
\end{tabular}
\label{tab:perp}
\vspace{-0.5cm}
\end{table*}

\vspace{-0.2cm}
\section{Experimental Setup}
\label{sec:exps}

\subsection{Databases}
\label{ssec:data}

\subsubsection{Spoken Material}
Details of all training data used for the experiments are presented in Table~\ref{tab:data}. The training data of the FAME! speech corpus comprises 8.5 hours and 3 hours of speech from Frisian and Dutch speakers respectively. The development and test sets consist of 1 hour of speech from Frisian speakers and 20 minutes of speech from Dutch speakers each. All speech data has a sampling frequency of 16 kHz.

The raw radio broadcast data extracted from the same archive as the FAME! speech corpus consists of 256.8 hours of audio, including 159.5 hours of speech based on the SAD~\cite{graciarena2016} output. The amount of total raw speech data extracted from the target broadcast archive after removing very short segments is 125.5 hours. We refer to this automatically annotated data as the `Frisian Broadcast' data.

Monolingual Dutch speech data comprises the complete Dutch and Flemish components of the Spoken Dutch Corpus (CGN)~\cite{cgn} that contains diverse speech material including conversations, interviews, lectures, debates, read speech and broadcast news. This corpus contains 442.5 and 307.5 hours of Dutch and Flemish data respectively.

\begin{table*}
\centering
\caption{WER (\%) obtained on the development and test set of the FAME! Corpus - Different AM training data is identified with the numbers which are defined in Table~\ref{tab:data}. The 3-fold data augmentation is applied for the numbers marked in bold. The numbers referring to different LM are defined in Table~\ref{tab:perp}. \text{``+R"} indicates applying rescoring using an RNN-LM trained on the corresponding text corpus.}
\addtolength{\tabcolsep}{-3.0pt}
\vspace{-0.2cm}
\begin{tabular}{| l | c | c || c c c c | c c c c | c |}
\hline
\multicolumn{3}{|c||}{} & \multicolumn{4}{c|}{Devel} & \multicolumn{4}{c|}{Test} & Total\\
\hline
\multicolumn{3}{|c||}{} & fy & nl & fy-nl & all & fy & nl & fy-nl & all & \\
\hline
 \multicolumn{3}{|c||}{\# of Frisian words} & 9190 & 0 & 2381 & 11,571 & 10,753 & 0 & 1798 & 12,551 & 24,122\\
\hline
 \multicolumn{3}{|c||}{\# of Dutch words} & 0 & 4569 & 533 & 5102 & 0 & 3475 & 306 & 3781 & 8883\\
\hline\hline
ASR System  & AM train data & LM & \multicolumn{4}{c|}{} & \multicolumn{4}{c|}{} & \\
\hline \hline
Baseline ASR  & (\bf{1}) & (1) & 36.4 & 43.7 & 48.2 & 40.3 & 31.5 & 39.5 & 47.9 & 35.2 & 37.8 \\
\hline
ASR\_AA       & (\textbf{1}+\textbf{2}) & (1) & 31.1 & 35.2 & 42.4 & 34.1 & 28.6 & 31.8 & 44.0 & 31.2 & 32.7 \\
\hline
ASR\_AA\_CGN-NL & (\textbf{1}+\textbf{2}+3) & (1) & 26.8 & 28.2 & 37.6 & 29.0 & 25.2 & 24.9 & 39.0 & 26.8 & 27.9 \\
\hline
ASR\_AA\_CGN-NL-VL & (\textbf{1}+\textbf{2}+3+4) & (1) & 26.3 & 27.6 & 36.8 & 28.4 & 25.1 & 24.4 & 39.3 & 26.7 & 27.6 \\
\hline
ASR\_AA\_CGN-NL++ & (\textbf{1}+\textbf{2}+\textbf{3}) & (1) & 25.8 & 27.4 & 36.9 & 28.1 & 24.8 & 24.6 & 37.6 & 26.4 & 27.2 \\
\hline
ASR\_AA\_CGN-NL-VL++ & (\textbf{1}+\textbf{2}+\textbf{3}+\textbf{4}) & (1) & 26.4 & 26.9 & 36.2 & 28.1 & 24.5 & 23.2 & 38.5 & 26.0 & 27.1 \\
\hline \hline
ASR\_AA\_CGN-NL-VL++\_RS & (\textbf{1}+\textbf{2}+\textbf{3}+\textbf{4}) & (1)+R & 24.3 & 25.2 & 36.2 & 26.5 & 22.9 & 21.8 & 36.5 & 24.3 & 25.4 \\
\hline
ASR\_AA\_CGN-NL-VL++\_CS-LM & (\textbf{1}+\textbf{2}+\textbf{3}+\textbf{4}) & (9) & 24.6 & 26.7 & 33.3 & 26.5 & 22.5 & 22.4 & 32.9 & 23.8 & 25.2 \\
\hline 
ASR\_AA\_CGN-NL-VL++\_CS-LM\_RS & (\textbf{1}+\textbf{2}+\textbf{3}+\textbf{4}) & (9)+R & \bf{22.6} & \bf{24.7} & \bf{31.4} & \bf{24.6} & \bf{21.2} & \bf{21.3} & \bf{30.1} & \bf{22.3} & \bf{23.5} \\
\hline 
\end{tabular}
\label{tab:wer}
\vspace{-0.5cm}
\end{table*}

\subsubsection{Written Material}

The baseline language models are trained on a bilingual text corpus containing 37M Frisian and 8.8M Dutch words. Almost all Frisian text is extracted from monolingual resources such as Frisian novels, news articles, Wikipedia articles. The Dutch text is extracted from the transcriptions of the CGN speech corpus which has been found to be very effective for language model training compared to other text extracted from written sources. The transcriptions of the FAME! training data is the only source of CS text and contains 140k words. 

Using this small amount of CS text, we train LSTM-LM and generate text with 10M, 25M, 50M and 75M words. The translated CS text contains 8.5M words. Finally, we use the automatic transcriptions provided by the best-performing monolingual and bilingual automatic transcription strategy which contains 3M words in total. The details of these strategies are described in~\cite{yilmaz2018}. The final training text corpus after including the generated text has 107M words.

\subsection{Implementation Details}
\label{ssec:impdet}

The recognition experiments are performed using the Kaldi ASR toolkit~\cite{kaldi}. We train a conventional context dependent Gaussian mixture model-hidden Markov model (GMM-HMM) system with 40k Gaussians using 39 dimensional mel-frequency cepstral coefficient (MFCC) features including the deltas and delta-deltas to obtain the alignments for training a lattice-free maximum mutual information (LF-MMI)~\cite{povey2016} TDNN-LSTM~\cite{peddinti2017} AM (1 standard, 6 time-delay and 3 LSTM layers) according to the standard recipe provided for the Switchboard database in the Kaldi toolkit (ver.~5.2.99). We use 40-dimensional MFCC combined with i-vectors for speaker adaptation~\cite{saon2013} and the default training parameters provided in the recipe without performing any parameter tuning. The 3-fold data augmentation~\cite{ko2015} is applied to the training data if mentioned. 

The baseline language models are standard bilingual 3-gram with interpolated Kneser-Ney smoothing and an RNN-LM~\cite{mikolov2010} with 400 hidden units used for recognition and lattice rescoring respectively. The RNN-LMs with gated recurrent units (GRU)~\cite{chung2014} and noise contrastive estimation~\cite{chen2015} are trained using the faster RNN-LM training implementation\footnote{https://github.com/yandex/faster-rnnlm}. A tied LSTM-LM~\cite{sundermeyer2012} with 650 hidden units per layer and 650-dimensional word embeddings is used for CS text generation which is trained for 40 epochs using the example Pytorch\footnote{https://github.com/pytorch/examples} implementation.

The bilingual lexicon contains 110k Frisian and Dutch words. The number of entries in the lexicon is approximately 160k due to the words with multiple phonetic transcriptions. The phonetic transcriptions of the words that do not appear in the initial lexicons are learned by applying grapheme-to-phoneme (G2P) bootstrapping~\cite{davel2003,maskey2004}. The lexicon learning is done only for the words that appear in the training data using the G2P model learned on the corresponding language. We use the Phonetisaurus G2P system~\cite{novak2015} for creating phonetic transcriptions.

\subsection{Perplexity and ASR Experiments}
\label{ssec:perpandrecexps}

We evaluate the quality of the language models by comparing the perplexities of the LMs trained on different components of the automatically generated text on the transcriptions of the development and test speech data. Later, the baseline and the LM with the lowest perplexity on the development transcriptions are used during the ASR experiments for recognition. The training text  corpus of both models are further used for training RNN-LMs for lattice rescoring.

We further run ASR experiments using various bilingual acoustic models trained on the setup summarized in Table~\ref{tab:data} on a separate set of manually annotated data used for testing purposes. The baseline acoustic models are trained only on the manually annotated data. Other ASR systems incorporate acoustic models trained on the combined (manually and automatically annotated) data and monolingual Dutch/Flemish data.

The evaluation is performed on the development and test data of the FAME! speech corpus and the recognition results are reported separately for Frisian only (fy), Dutch only (nl) and mixed (fy-nl) segments. The overall performance (all) is also provided as a performance indicator. The recognition performance of the ASR system is quantified using the word error rate (WER). The word language tags are removed while evaluating the ASR performance.

\section{Results}
\label{sec:res}

\subsection{Perplexity Results}
\label{ssec:perpres}
The perplexity results obtained on the development and test set transcriptions of the FAME! speech database are given in Table~\ref{tab:perp}. The baseline language model has a perplexity of 257.9 and 231.6 on the development and test set respectively. Firstly, various amounts of CS text generated using an LSTM-LM are added to the training text corpus. All LMs trained on the baseline and LSTM-LM generated text, namely (2)-(5), provide considerably lower perplexities on both the development and test transcriptions. Adding 50M words provides the lowest perplexity (195.2) on the development set, thus, we use this combination for training the the following models. 

Using the automatically generated transcriptions hypothesized by two different automatic transcription strategies results in further perplexity reduction. Adding each set of transcriptions bring similar improvements by reducing the perplexity to 188 and 166 on the development and test sets respectively. Merging both corpora reduces the perplexity to 179.8 on the development text and 157.9 on the test text. Finally, including the machine translated text brings marginal improvements on both sets with a final perplexity of 178.1 on the development and 155.8 on the test set.

\subsection{ASR Results}
\label{ssec:asrres}
The ASR results obtained on the development and test sets of the FAME! speech database are presented in Table~\ref{tab:wer}. The number of Frisian and Dutch words in the development and test sets are given in the upper panel. The baseline ASR trained only on the available CS training data of 11.5 hours provides a total WER of 37.8\%. Due to the increase training data due to automatic annotation strategies, the total WER reduces to 32.7\%. For both the baseline ASR and ASR\_AA, the data augmentation is applied during training as only in-domain training data is used in these cases.

In the next step, we add a large amount of monolingual Dutch data without applying data augmentation in order to avoid adverse effects of data imbalance between the mixed languages. This results in a further decrease in 4.8\% absolute improvement in the total WER. Unlike reported in previous work\cite{yilmaz2016_4}, adding monolingual Dutch does not hinder the recognition of Frisian words after the increase (from {\raise.17ex\hbox{$\scriptstyle\sim$}}10 hours to {\raise.17ex\hbox{$\scriptstyle\sim$}}135 hours) in the in-domain CS training data. 

Further adding the Flemish component brings marginal improvements in the ASR performance with a WER of 27.6\%. Applying data augmentation to the Dutch and Flemish monolingual data (ASR\_AA\_CGN-NL-VL++) reduces the total WER to 27.1\%. Adding Flemish mildly improving the recognition of both Frisian and Dutch utterances, it is difficult to conclude if the improvements are because of Flemish helping with the Frisian-accented Dutch or simply due to the increase in the amount of training speech material. 

After using all available acoustic data for AM training, we proceed with exploring the impact of the enriched CS LM and RNN-LM rescoring on the ASR performance. As expected, the enriched CS LM mostly helps with the recognition of the mixed (fy-nl) segments by reducing the WER from 36.2\% to 33.3\% on the development and from 38.5\% to 32.9\% on the test set. Applying lattice rescoring using RNN-LMs trained on both the baseline and enriched text corpus yields an absolute improvement of 1.7\% compared to the corresponding ASR system without rescoring. The best performing ASR\_AA\_CGN-NL-VL++\_CS-LM\_RS system has a total WER of 23.5\% with a lower WER on all components of the development and test segments compared to the previous ASR systems.

\section{Conclusions}
\label{sec:conc}

In this work, we describe several techniques to improve the acoustic and language modeling of a CS ASR system. Exploiting monolingual speech data from the high-resourced mixed language for improving the AM quality is found to be viable after increasing the amount of in-domain speech, for instance, by performing automatic transcription of raw data resembling the target speech. Moreover, increasing the amount of CS text by text generation using recurrent LMs trained on a very small amount of reference CS text and automatic transcriptions from different transcription strategies has provided enriched LMs that has significantly lower perplexities on the development and test transcriptions. These enriched LMs have also reduced the WER especially on the mixed segments containing words from both languages.   

\bibliographystyle{IEEEtran}

\bibliography{refs}

\begin{thebibliography}{10}
\providecommand{\url}[1]{#1}
\csname url@samestyle\endcsname
\providecommand{\newblock}{\relax}
\providecommand{\bibinfo}[2]{#2}
\providecommand{\BIBentrySTDinterwordspacing}{\spaceskip=0pt\relax}
\providecommand{\BIBentryALTinterwordstretchfactor}{4}
\providecommand{\BIBentryALTinterwordspacing}{\spaceskip=\fontdimen2\font plus
\BIBentryALTinterwordstretchfactor\fontdimen3\font minus
  \fontdimen4\font\relax}
\providecommand{\BIBforeignlanguage}[2]{{%
\expandafter\ifx\csname l@#1\endcsname\relax
\typeout{** WARNING: IEEEtran.bst: No hyphenation pattern has been}%
\typeout{** loaded for the language `#1'. Using the pattern for}%
\typeout{** the default language instead.}%
\else
\language=\csname l@#1\endcsname
\fi
#2}}
\providecommand{\BIBdecl}{\relax}
\BIBdecl

\bibitem{stemmer2001}
G.~Stemmer, E.~N{\"{o}}th, and H.~Niemann, ``Acoustic modeling of foreign words
  in a {German} speech recognition system,'' in \emph{Proc. EUROSPEECH}, 2001,
  pp. 2745--2748.

\bibitem{lyu2006}
D.-C. Lyu, R.-Y. Lyu, Y.-C. Chiang, and C.-N. Hsu, ``Speech recognition on
  code-switching among the {Chinese} dialects,'' in \emph{Proc. ICASSP},
  vol.~1, May 2006, pp. 1105--1108.

\bibitem{vu2012}
N.~T. Vu, D.-C. Lyu, J.~Weiner, D.~Telaar, T.~Schlippe, F.~Blaicher, E.-S.
  Chng, T.~Schultz, and H.~Li, ``A first speech recognition system for
  {Mandarin-English} code-switch conversational speech,'' in \emph{Proc.
  ICASSP}, March 2012, pp. 4889--4892.

\bibitem{modipa2013}
T.~I. Modipa, M.~H. Davel, and F.~De~Wet, ``Implications of {Sepedi/English}
  code switching for {ASR} systems,'' in \emph{Pattern Recognition Association
  of South Africa}, 2015, pp. 112--117.

\bibitem{lyudovyk2014}
T.~Lyudovyk and V.~Pylypenko, ``Code-switching speech recognition for closely
  related languages,'' in \emph{Proc. SLTU}, 2014, pp. 188--193.

\bibitem{wu2014}
C.~H. Wu, H.~P. Shen, and C.~S. Hsu, ``Code-switching event detection by using
  a latent language space model and the delta-{Bayesian} information
  criterion,'' \emph{IEEE/ACM Transactions on Audio, Speech, and Language
  Processing}, vol.~23, no.~11, pp. 1892--1903, Nov 2015.

\bibitem{yilmaz2016_2}
E.~Y{\i}lmaz, H.~Van~den Heuvel, and D.~A. Van~Leeuwen, ``Investigating
  bilingual deep neural networks for automatic speech recognition of
  code-switching {Frisian} speech,'' in \emph{Proc. SLTU}, May 2016, pp.
  159--166.

\bibitem{weiner2012}
J.~Weiner, N.~T. Vu, D.~Telaar, F.~Metze, T.~Schultz, D.-C. Lyu, E.-S. Chng,
  and H.~Li, ``Integration of language identification into a recognition system
  for spoken conversations containing code-switches,'' in \emph{Proc. SLTU},
  May 2012.

\bibitem{lyu2013}
D.-C. Lyu, E.-S. Chng, and H.~Li, ``Language diarization for code-switch
  conversational speech,'' in \emph{Proc. ICASSP}, May 2013, pp. 7314--7318.

\bibitem{yeong2014}
Y.-L. Yeong and T.-P. Tan, ``Language identification of code switching
  sentences and multilingual sentences of under-resourced languages by using
  multi structural word information,'' in \emph{Proc. INTERSPEECH}, Sept. 2014,
  pp. 3052--3055.

\bibitem{mabokela2014}
K.~R. Mabokela, M.~J. Manamela, and M.~Manaileng, ``Modeling code-switching
  speech on under-resourced languages for language identification,'' in
  \emph{Proc. SLTU}, 2014, pp. 225--230.

\bibitem{li2012}
Y.~Li and P.~Fung, ``Code switching language model with translation constraint
  for mixed language speech recognition,'' in \emph{Proc. COLING}, Dec. 2012,
  pp. 1671--1680.

\bibitem{adel2013}
H.~Adel, N.~Vu, F.~Kraus, T.~Schlippe, H.~Li, and T.~Schultz, ``Recurrent
  neural network language modeling for code switching conversational speech,''
  in \emph{Proc. ICASSP}, 2013, pp. 8411--8415.

\bibitem{adel2014}
H.~Adel, K.~Kirchhoff, D.~Telaar, N.~T. Vu, T.~Schlippe, and T.~Schultz,
  ``Features for factored language models for code-switching speech,'' in
  \emph{Proc. SLTU}, May 2014, pp. 32--38.

\bibitem{zeng2017}
Z.~Zeng, H.~Xu, T.~Y. Chong, E.-S. Chng, and H.~Li, ``Improving {N-gram}
  language modeling for code-switching speech recognition,'' in \emph{Proc.
  APSIPA ASC}, 2017, pp. 1--6.

\bibitem{hamed2017}
I.~Hamed, M.~Elmahdy, and S.~Abdennadher, ``Building a first language model for
  code-switch {Arabic-English},'' \emph{Procedia Computer Science}, vol. 117,
  pp. 208 -- 216, 2017.

\bibitem{westhuizen2017}
E.~van~der Westhuizen and T.~Niesler, ``Synthesising {isiZulu-English}
  code-switch bigrams using word embeddings,'' in \emph{Proc. INTERSPEECH},
  2017, pp. 72--76.

\bibitem{yilmaz2016_4}
E.~Y{\i}lmaz, H.~van~den Heuvel, and D.~van Leeuwen, ``Code-switching detection
  using multilingual {DNN}s,'' in \emph{IEEE Spoken Language Technology
  Workshop (SLT)}, Dec 2016, pp. 610--616.

\bibitem{yilmaz2017_2}
E.~Y{\i}lmaz, M.~McLaren, H.~Van~den Heuvel, and D.~A. Van~Leeuwen, ``Language
  diarization for semi-supervised bilingual acoustic model training,'' in
  \emph{Proc. ASRU}, Dec. 2017, pp. 91--96.

\bibitem{yilmaz2018}
\BIBentryALTinterwordspacing
------, ``Semi-supervised acoustic model training for speech with
  code-switching,'' \emph{Submitted to Speech Communication}, 2018. [Online].
  Available: \url{goo.gl/NKiYAF}
\BIBentrySTDinterwordspacing

\bibitem{peddinti2017}
V.~Peddinti, Y.~Wang, D.~Povey, and S.~Khudanpur, ``Low latency acoustic
  modeling using temporal convolution and {LSTM}s,'' \emph{IEEE Signal
  Processing Letters}, vol.~25, no.~3, pp. 373--377, March 2018.

\bibitem{yilmaz2016}
E.~Y{\i}lmaz, M.~Andringa, S.~Kingma, F.~Van~der Kuip, H.~Van~de Velde,
  F.~Kampstra, J.~Algra, H.~Van~den Heuvel, and D.~Van~Leeuwen, ``A
  longitudinal bilingual {Frisian-Dutch} radio broadcast database designed for
  code-switching research,'' in \emph{Proc. LREC}, 2016, pp. 4666--4669.

\bibitem{sutskever2011}
I.~Sutskever, J.~Martens, and G.~Hinton, ``Generating text with recurrent
  neural networks,'' in \emph{Proc. ICML-11}, June 2011, pp. 1017--1024.

\bibitem{provinciefryslan}
{Provinsje Frysl\^{a}n}, ``{De Fryske taalatlas 2015. De Fryske taal yn
  byld},'' 2015, available at http://www.fryslan.frl/taalatlas.

\bibitem{myers1989}
C.~Myers-Scotton, ``Codeswitching with {English}: types of switching, types of
  communities,'' \emph{World Englishes}, vol.~8, no.~3, pp. 333--346, 1989.

\bibitem{kaldi}
D.~Povey, A.~Ghoshal, G.~Boulianne, L.~Burget, O.~Glembek, N.~Goel,
  M.~Hannemann, P.~Motlicek, Y.~Qian, P.~Schwarz, J.~Silovsky, G.~Stemmer, and
  K.~Vesely, ``The {Kaldi} speech recognition toolkit,'' in \emph{Proc. ASRU},
  Dec. 2011.

\bibitem{bezooijen1999}
R.~v. Bezooijen and J.~Ytsma, ``Accents of {Dutch}: personality impression,
  divergence, and identifiability,'' \emph{Belgian Journal of Linguistics}, pp.
  105--–129, 1999.

\bibitem{jensson2009}
A.~Jensson, K.~Iwano, and S.~Furui, ``Language model adaptation using
  machine-translated text for resource-deficient languages,'' \emph{EURASIP
  Journal on Audio, Speech, and Music Processing}, vol. 2008, no.~1, pp. 1--7,
  Jan 2009.

\bibitem{cgn}
N.~Oostdijk, ``The spoken {Dutch} corpus: {Overview} and first evaluation,'' in
  \emph{Proc. LREC}, 2000, pp. 886--894.

\bibitem{graciarena2016}
M.~Graciarena, L.~Ferrer, and V.~Mitra, ``The {SRI System for the NIST OpenSAD}
  2015 speech activity detection evaluation,'' in \emph{Proc. INTERSPEECH},
  2016, pp. 3673--3677.

\bibitem{povey2016}
D.~Povey, V.~Peddinti, D.~Galvez, P.~Ghahremani, V.~Manohar, X.~Na, Y.~Wang,
  and S.~Khudanpur, ``Purely sequence-trained neural networks for {ASR} based
  on lattice-free {MMI},'' in \emph{Proc. INTERSPEECH}, 2016, pp. 2751--2755.

\bibitem{saon2013}
G.~Saon, H.~Soltau, D.~Nahamoo, and M.~Picheny, ``Speaker adaptation of neural
  network acoustic models using i-vectors,'' in \emph{Proc. ASRU}, Dec 2013,
  pp. 55--59.

\bibitem{ko2015}
T.~Ko, V.~Peddinti, D.~Povey, and S.~Khudanpur, ``Audio augmentation for speech
  recognition,'' in \emph{Proc. INTERSPEECH}, 2015, pp. 3586--3589.

\bibitem{mikolov2010}
T.~Mikolov, M.~Karafi{\'{a}}t, L.~Burget, J.~Cernock{\'{y}}, and S.~Khudanpur,
  ``Recurrent neural network based language model,'' in \emph{Proc.
  INTERSPEECH}, 2010, pp. 1045--1048.

\bibitem{chung2014}
J.~Chung, C.~Gulcehre, and Y.~Kyung Hyun~Cho, Bengio, ``Empirical evaluation of
  gated recurrent neural networks on sequence modeling,''
  \emph{arXiv:1412.3555}, 2014.

\bibitem{chen2015}
X.~Chen, X.~Liu, M.~J.~F. Gales, and P.~C. Woodland, ``Recurrent neural network
  language model training with noise contrastive estimation for speech
  recognition,'' in \emph{Proc. ICASSP}, April 2015, pp. 5411--5415.

\bibitem{sundermeyer2012}
M.~Sundermeyer, R.~Schl{\"u}ter, and H.~Ney, ``{LSTM} neural networks for
  language modeling,'' in \emph{Proc. INTERSPEECH}, 2012, pp. 194--197.

\bibitem{davel2003}
M.~Davel and E.~Barnard, ``Bootstrapping for language resource generation,'' in
  \emph{Pattern Recognition Association of South Africa}, 2003, pp. 97--100.

\bibitem{maskey2004}
S.~R. Maskey, A.~B. Black, and L.~M. Tomokiyo, ``Bootstrapping phonetic
  lexicons for new languages,'' in \emph{Proc. ICLSP}, 2004, pp. 69--72.

\bibitem{novak2015}
J.~R. Novak, N.~Minematsu, and K.~Hirose, ``Phonetisaurus: Exploring
  grapheme-to-phoneme conversion with joint n-gram models in the {WFST}
  framework,'' \emph{Natural Language Engineering}, pp. 1--32, 9 2015.

\end{thebibliography}

\end{document}